# Structured Knowledge Base Enhances Effective Use of Large Language Models for Metadata Curation

Sowmya S. Sundaram, Ph.D.[1], Benjamin Solomon, M.D., Ph.D.[1,2], Avani Khatri M.S.[2], Anisha Laumas A.B.[1,2], Purvesh Khatri, Ph.D.[1,2] and Mark A. Musen, M.D., Ph.D.[1]

[1]Center for Biomedical Informatics Research, School of Medicine, Stanford University, Stanford, California, USA; [2]Institute for Immunity, Transplantation and Infection, School of Medicine, Stanford University, Stanford, California, USA

**Abstract**

*Metadata play a crucial role in ensuring the findability, accessibility, interoperability, and reusability of datasets. This paper investigates the potential of large language models (LLMs), specifically GPT-4, to improve adherence to metadata standards in existing datasets. We conducted experiments on 200 random data records describing human samples relating to lung cancer from the NCBI BioSample repository, evaluating GPT-4's ability to suggest edits for adherence to metadata standards. We computed the adherence accuracy of field name–field value pairs through a peer review process, and we observed a marginal average improvement in adherence to the standard data dictionary from 79% to 80% when using GPT-4. We then prompted GPT-4 with domain information in the form of the textual descriptions of CEDAR metadata templates and recorded a statistically significant improvement to 97% from 79% (p<0.01). These results indicate that, while LLMs show promise for use in automated metadata curation when integrated with a structured knowledge base, though they may struggle when unaided.*

**Introduction**

Data sharing, a pivotal requirement now required by most funding agencies, continues to be a challenging prospect. Researchers hoping to take advantage of shared datasets in public repositories encounter many roadblocks, such as finding relevant datasets, understanding precisely what the original investigators have done, and comparing myriad sources. The desire to share data in a manner that promotes findability, accessibility, interoperability, and reusability led to the articulation of the FAIR principles[1]. The FAIR principles emphasize the importance of *metadata* (annotations that describe the corresponding dataset and the experiments that were performed to obtain the data) and of metadata that are "rich" and that adhere to the standards of the relevant scientific community[2]. We propose an automated method of performing metadata standardization using natural language processing techniques and a metadata knowledge base.

Creating metadata standards and evaluating adherence to these standards are difficult, due to several factors. The issue of *representational heterogeneity*, for example, is one major roadblock in ensuring adherence to metadata standards. There are plenty of ways by which metadata can be represented (for example, one can label age as "age," "age in years," "Age," and so on). This heterogeneity has led many scientific communities to adopt metadata standards. Unfortunately, ignorance of these standards—or the existence of multiple standards—reduces their efficacy. More often, investigators simply choose to ignore standards. Our prior research on BioSample[3] records reveals significant discrepancies in the adherence of metadata to the data dictionary (as specified by NCBI), such as the widespread absence of adherence to data types, failure to use recommended names for metadata fields, and disregard for suggested ontologies when entering field values[4].

The task of evaluating and correcting metadata to adhere to community standards involves language understanding. Since the advent of Large Language Models (LLMs), many attempts have been made to exploit the language understanding capabilities of LLMs for a wide variety of domains[4,5]. The large number of trainable parameters, coupled with the enormous amounts of text data available on the World Wide Web used to train LLMs, leads LLMs to model language well. Some applications of LLMs described in the literature are aimed at metadata extraction and generation[6,7]. For example, missing metadata might be filled by *prompting* the LLM to summarize data samples.

One way of extracting information from LLMs is by prompt engineering. Prompt engineering is a commonly used technique for eliciting domain-specific information. Prompts are helper texts added to queries that users make to LLMs, such that the helper texts resemble text on which the LLMs were trained and can help to refine the LLM's response. One common method of prompting involves describing how a typical input and output should look, guiding the model's response to match expected formats or styles. For metadata adherence, we may add instructions to clarify the structure of the desired metadata. Another powerful prompting method is known as *few-shot prompting*. In this approach, the prompt includes a few examples of the desired input–output pairs. For metadata adherence, few shot prompting could include pairs of "incorrect" and "desired" metadata. By providing these examples, the model is better able to infer the task at hand and to generate appropriate responses, even when the prompt is new or unfamiliar to the model. In many of these applications, the modus operandi is to utilize prompts to generate metadata descriptions from datasets for downstream tasks such as question answering and data-record classification.

Another method of elicitng information stored in LLMs is to use additional domain knowledge in the input[10,11]. In our work, we leverage computer-stored domain knowledge to encourage adherence to metadata standards. The CEDAR Workbench is a platform designed to facilitate the creation and sharing of metadata for scientific datasets.[9] CEDAR offers a Web-based interface that allows users to encode community standards for metadata in a machine-actionable form. Hence, CEDAR can provide a knowledge base for metadata. Figure 1 presents the CEDAR metadata template for BioSample and shows how the system suggests values for the field named "tissue."

To our knowledge, our work is the first effort using LLMs for metadata correction with a rigorous emphasis on field values adhering to ontological restrictions if applicable, and with a peer evaluation that establishes the usefulness of our approach. This work builds on the research previously done by our laboratory in exploring word embeddings for metadata correction[8]. In this paper, we explore how LLMs can correct the metadata used to annotate legacy datasets to improve adherence to metadata guidelines. We first use LLMs by themselves to correct existing metadata in a public dataset. We conducted our initial experiments on the BioSample database, which we selected due to its popularity, extensive metadata descriptions, and our laboratory's prior familiarity with it. With prompt engineering, we assessed the ability of the popular LLM GPT-4 to make metadata adhere to BioSample's metadata standards. In our experiment, we found unsatisfactory error rates (nearly 20%) when GPT-4 acted unaided, but these rates decreased significantly (to 3%,) when we enhanced GPT-4's performance by providing it with access to a structured knowledge base in the form of CEDAR metadata templates.

**Methods**
We selected a conveneience dataset consisting of 200 records from BioSample using the Mersenne Twister algorithm for randomization[12]. We selected records that were associated with human samples that contained a mention of "lung cancer."  We chose to explore lung cancer as our experimental domain due to the abundance of varied metadata descriptions within a compact sample size. We then chose GPT-4 as the LLM for our work, as it has been reported to perform competitively on a wide range of tasks[13]. Our goal was to assess the metadata's adherence to community standards across three versions of each metadata record:
1. The original BioSample record (hereinafter referred to as BioSample)
2. GPT-4's correction of the original record without reference to the CEDAR template (hereinafter referred to as LLM)
3. GPT-4's correction with the CEDAR template as input (hereinafter referred to as LLM+CEDAR)

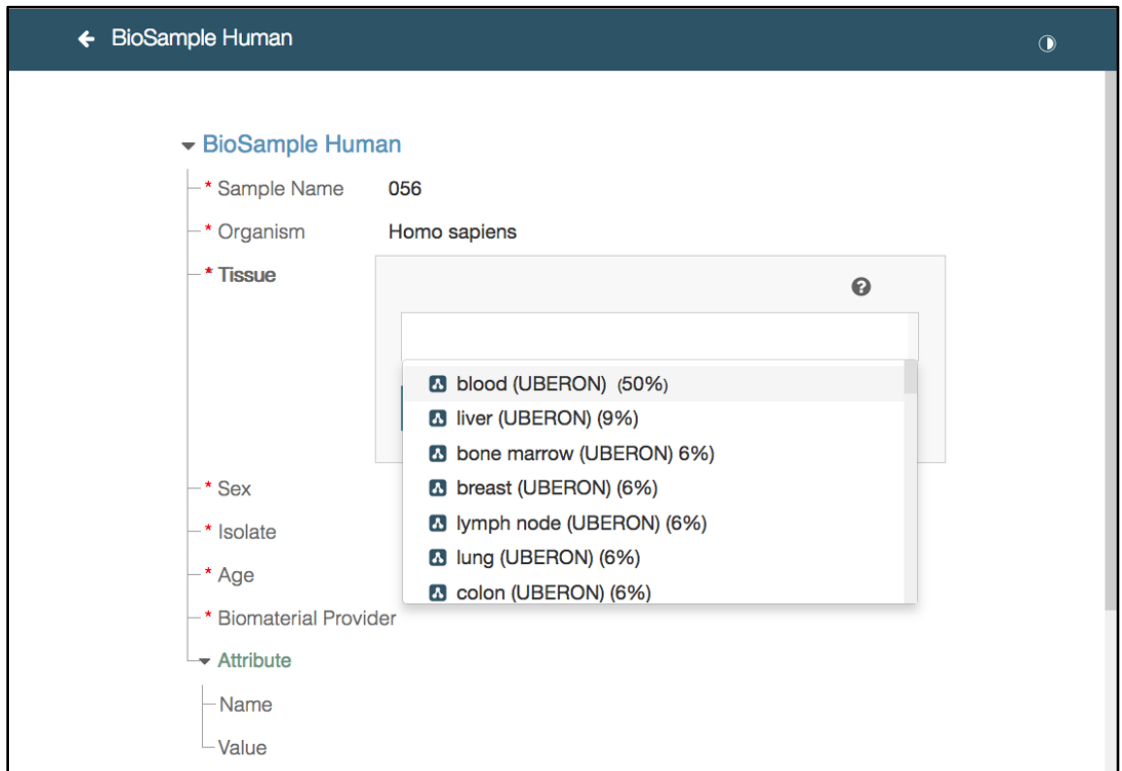

**Figure 1.** Screen capture of the CEDAR metadata template for BioSample. The template consists of a set of standard attributes (or field names), on the left. The user is creating an instance of metadata by supplying values for the attributes, on the right. The field "tissue" takes on values from a branch on the UBERON[14] anatomy ontology. Thus, when the user clicks on this field, CEDAR displays possible values from the ontology, sorted in order of likelihood based on analysis of previous metadata entries for this template.

For the LLM setting, we provided GPT-4 with a prompt containing a string that introduced the text of the BioSample record and we outlined the desired attributes of metadata quality. The prompt we used was, "Here is a record from BioSample describing a sample from a patient diagnosed with lung cancer. Ensure that the field names and field values make sense." For the LLM+CEDAR setting, we provided GPT-4 with a prompt additionally containing the template provided by CEDAR for BioSample. The prompt was: "Here is a record from BioSample describing a sample from a patient diagnosed with lung cancer. Ensure that the field names and field values make sense. Following the BioSample record is a template describing the allowed field names and values. Transform the record to adhere to the template."

We used two methods of evaluation to assess the adherence of our test metadata to community standards: (1) automated evaluation and (2) peer evaluation. For **automated evaluation**, we wrote a program that tested whether field values adhered to ontological restrictions for certain field names. We chose three field names that were linked to ontologies in the BioSample data dictionary. For the evaluation, we utilized the outcome measure of determining the percentage of fields in both the original and "corrected" metadata data whose values adhered to the specified ontological restriction. This matching was performed through string matching to a list of class names. The fields considered were (1) *tissue* (matching to UBERON), (2) *disease* (matching to the Disease Ontology), and (3) *cell type* (matching to the Cell Ontology). For **peer evaluation**, three of us (a pediatrics research postdoctoral scholar [BS], an immunology researcher [AL], and a medical data scientist [AK]) worked with sample datasets to identify the errors in the data records. In the peer evaluation setup, we recorded two measures for every record that we considered from BioSample:
1. Adherence accuracy of a record = 1 - (number of adherence errors per field / total number of fields)
2. Error count of a record = the total number of errors per record

| BioSample Record (Original) | GPT-4 "Correction" (LLM) | LLM + CEDAR |
|---|---|---|
| isolate : TN_32<br>age : 67<br>sex : female<br>**tissue : lung cancer** | Sample Isolate Identifier: TN_32<br>Age of Donor: 67<br>Sex of Donor: Female<br>**Sampled Tissue: Lung (afflicted with cancer)** | biosample_accession: NA<br>sample_name: TN_32<br>bioproject_accession: NA<br>**organism: Homo sapiens**<br>isolate: TN_32<br>age: 67<br>sex: female<br>**tissue: lung**<br>**disease: lung cancer**<br>**health_state: Diseased**<br>treatment: NA<br>ethnicity: NA |

**Figure 2:** LLM correction for metadata adherence with the CEDAR template added to the prompt. On the left we see a portion of the original BioSample record. In the middle is the GPT-4 "correction" of the erroneous metadata. Although "lung cancer" is not a type of tissue, GPT-4 hallucinates when "correcting" the entry to "lung (afflicted with cancer)." On the right is the revised metadata record when the CEDAR template for BioSample is added to the GPT-4 prompt. The resulting record is more complete, and the metadata values adhere to the standard defined by the template. The boldface (added) highlights the error in the original BioSample record and the attempts to correct it.

A pertinent point to note is that the automated evaluation and the peer evaluation concern two different tasks. In the automated evaluation, adherence accuracy of specific field–value pairs are recorded. In the peer evaluation, every field is evaluated and aggregated according to the formulae presented above. In automated evaluation, we choose three specific field names that are usually erroneous in legacy metadata and for which it is hard to ensure correct value restrictions. For example, one can easily check whether the field value for age is an integer between 0–120. Checking adherence to a list of ontological values is significantly harder in the absence of an automatd system. In peer evaluation, the reviewers examined *all* fields in the metadata records;hence, the resulting measures are different.

**Results**

First, we begin with examples of our method. We first examine a sample from BioSample. Figure 2 shows an example of a metadata record with an obvious error, before (left column) and after being "corrected" by GPT-4 (middle column). GPT-4 appropriately recognized that "lung cancer" is not a tissue and it attempted to correct it. However, without access to the metadata guidelines, it did not ensure that the field value should be an entry from "Uberon Multi-Species Anatomy Ontology (UBERON)[14]," for better findability. The right-hand column in Figure 2 shows how the record was corrected by GPT-4 working in conjunction with the BioSample template from the CEDAR Workbench.

The results of the automated evaluation are presented in Figure 3. Consider the BioSample record in Figure 2 which has 1 error (the "tissue" field name has a value that corresponds to a disease). The error count for this record is 1 and the adherence accuracy of the record is 0.75, as 3 field names out of 4 have no obvious error. We calculate the error count irrespective of the number of field names to demonstrate that, despite the results of "LLM+CEDAR" having more field names introduced by consideration of the template, the mean error is reduced. On average, the adherence to standards for LLM+CEDAR-informed records was ~40% higher than that of the original BioSample records ($p < 0.01$). (We used *t*-tests for all the statistical significance calculations, as the *t*-test is best suited for proportional values[15].) Figures 4 and 5 show the results of peer evaluation. On average (the last set of bars), the correctness of "LLM" is minimally better than "BioSample" ($p=0.2$) and the correctness of "LLM+CEDAR" is significantly better than "BioSample" ($p<0.01$). Figure 5 presents the mean error count. Here also, on average (again the last set of bars), the error count of "LLM" is less than that of "BioSample" ($p=0.2$) and the error count of "LLM+CEDAR" is less than that of "LLM" ($p<0.01$). Since the peer evaluation is examining more fields than in the case of the automated evaluation, one may expect the peer evaluation to detect more errors. However, the automated evaluation examined the most error-prone field–value pairs that we identified a prioi (the ones that require adherence to ontological concepts), and hence the automated evaluation was a challenging task. With peer evaluation, the reviewers were additonally considering medical errors whose detection cannot be automated. Hence peer evaluation is a superior, although time-consuming, process. It was the onerous nature of the peer-evaluation process that caused us to limit our sample size to 200 BioSample records.

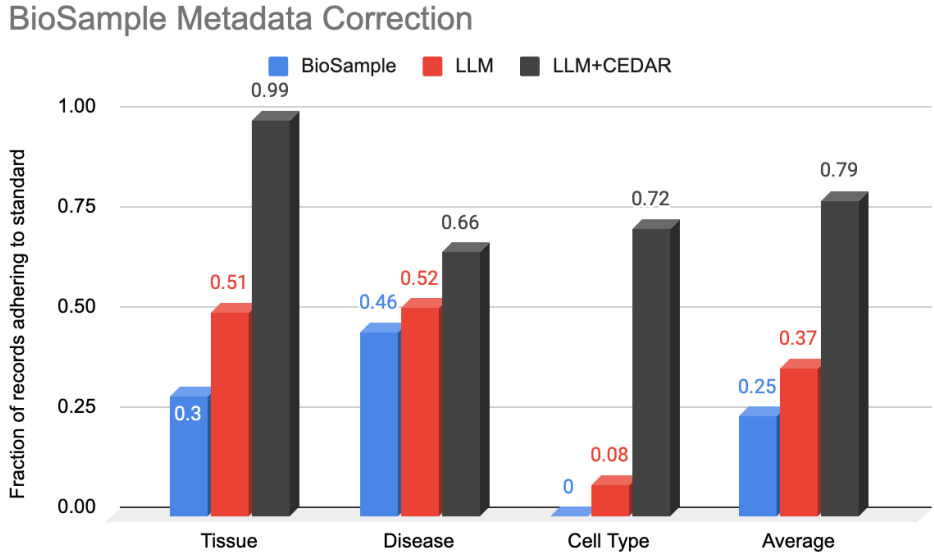

**Figure 3.** Mean adherence accuracy of three fields– 'tissue', 'disease' and 'cell type'. For each field in each record, we checked whether the corresponding value adhered to the ontological restriction recommended by the BioSample data dictionary (tissue: UBERON, disease: Disease Ontology, and cell type: Cell Ontology). The fraction of adherent values to the total number of such values in all records is presented in this figure for the three settings – BioSample, LLM, and LLM+CEDAR

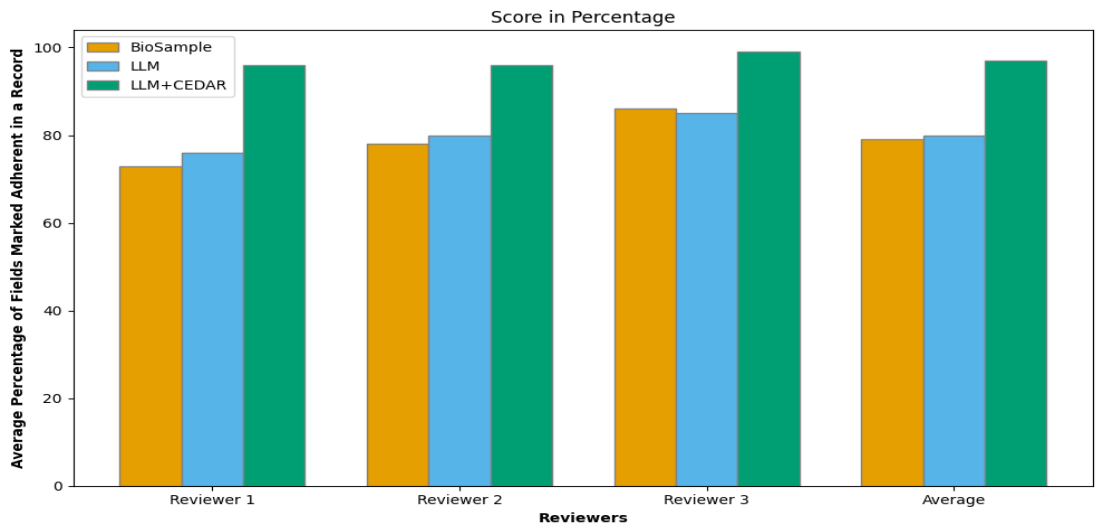

**Figure 4.** The accuracy scores given by three reviewers to the different versions of the records: BioSample, LLM, and LLM+CEDAR. Every field–value pair is evaluated for adherence to standard and the *percentage of correct pairs to the total* is averaged over the 200 records. The average score across all reviewers, for each type, is shown on the rightmost set of bars. On average, LLM+CEDAR records were evaluated to score better.

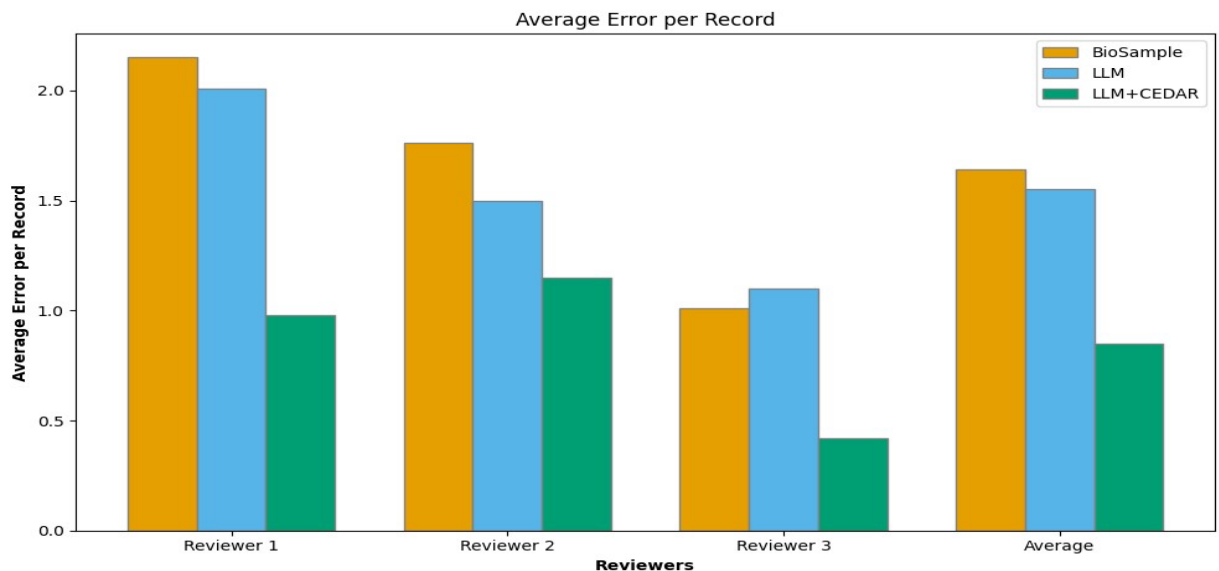

**Figure 5.** The number of errors recorded by the three reviewers for the different versions of the records: BioSample, LLM, and LLM+CEDAR. Every field–value pair was evaluated for consistency and adherence and the *number of errors* was recorded. Then, these error values were averaged over the 200 records. The average score across all reviewers, for each type, is shown on the rightmost set of bars. On average, LLM+CEDAR records have the fewest errors, despite having more fields per record.

**Table 1.** Inter-rater agreement on record errors among peer reviewers[1]

|  | Reviewer 1 | Reviewer 2 | Reviewer 3 |
| --- | --- | --- | --- |
| Reviewer 1 | 1.00 |  |  |
| Reviewer 2 | 0.34 | 1.00 |  |
| Reviewer 3 | 0.33 | 0.46 | 1.00 |

In Table 1, we present the inter-rater agreement of adherence errors in records among the three peer reviewers for all the three versions of the samples, using Kendall's Tau measure[16]. The measure captures trends in agreement.

The peer evaluation makes the usefulness of adding CEDAR templates to GPT-4 prompts clear. Generally, the LLM+CEDAR records exhibit notably better adherence to standard than do the original BioSample records. When adding the CEDAR metadata template to the GPT-4 prompt, the average correctness increases from 79 percent to 97 percent ($p<0.01$) and the average error reduces from 1.64 per record to 0.85 per record ($p<0.01$). This result is especially interesting, as the LLM+CEDAR version, on average, has more field names than the original BioSample record.

---

[1] The rows and columns depict the grid of reviewers and the pair-wise reviewer agreement according to Kendall's Tau.

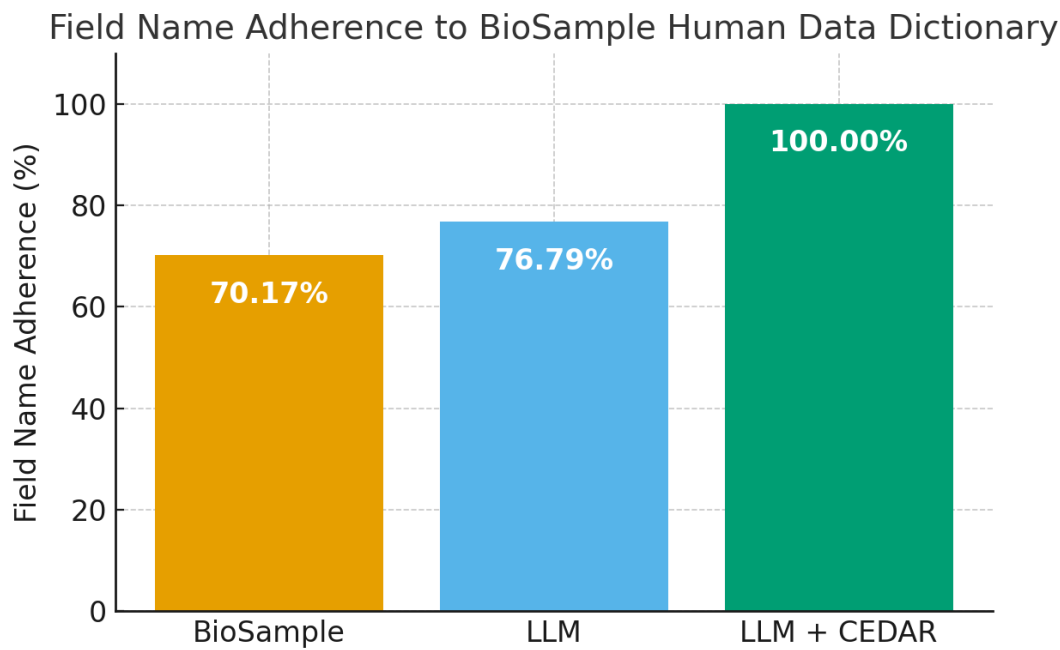

**Figure 6**. Field Name Adherence to BioSample Human Data Dictionary. The adherence percentage represents the proportion of field names that match the BioSample human data dictionary across three settings—BioSample (original records), LLM (GPT-4 corrections), and LLM+CEDAR (GPT-4 corrections with CEDAR metadata templates). The LLM+CEDAR approach achieved perfect adherence, highlighting the role of structured metadata knowledge in improving standardization.

In addition to the previous evaluation of metadata adherence, we conducted an experiment focusing specifically on field name adherence to the BioSample package human data dictionary. This experiment assessed how well field names in BioSample records conformed to the expected standard. We compared three conditions: the original BioSample records, records corrected using GPT-4 (LLM), and records corrected using GPT-4 with CEDAR metadata templates (LLM+CEDAR). As shown in Figure 6, the original BioSample records exhibited 70.17% adherence to the data dictionary, while the LLM-enhanced version showed a marginal improvement to 76.79%. However, when CEDAR templates were incorporated into the LLM corrections, adherence reached 100%, demonstrating the effectiveness of structured metadata guidance in ensuring compliance with field name conventions.

**Discussion**

The introduction of LLMs has sparked significant excitement and anticipation in many scientific disciplines and in the general population. The large size of LLM training data and the wide scope of potential subject matter has led to massive experimentation. However, this enthusiasm is often met with disappointment when LLMs are utilized in contexts that are highly specialized, require high precision, or require knowledge that is not easily describable through text. This discrepancy between expectation and reality highlights a critical challenge in effectively deploying LLMs across diverse applications. In this paper, we initiate our discussion by acknowledging this phenomenon, recognizing the superior language modeling skills of the LLMs and, at the same time, highlight the need for caution for many applications and domains. By addressing this issue head-on, we aim to contribute to a deeper understanding of the practical considerations and challenges associated with deploying LLMs in real-world settings.

The idea of including domain knowledge for enhancing the performance of large language models is an established doctrine. Traditionally, incorporating domain knowledge into language models has been achieved through methods such as fine-tuning with domain-specific examples or employing specialized language models trained on domain-specific training data, such as BioMedLM[17]. However, these approaches face limitations for specialized domains and complicated downstream processing tasks. Some of these limitations include having access to a large number of high quality hand-crafted input-output pairs and having access to a large body of text on metadata (which is not readily available).

Metadata, while crucial for structuring and contextualizing data, present unique obstacles for language models such as GPT-4. LLMs such as GPT-4 are trained using vast amounts of textual data from diverse sources such as books, articles, and websites. This process, known as pre-training, involves feeding the model with this extensive dataset to learn the structure, grammar, context, and nuances of the language. During pre-training, the model learns to predict the next word in a sentence, thereby acquiring a broad representation of linguistic patterns and knowledge. However, one of the critical challenges in developing LLMs for specific tasks, such as metadata curation, is the scarcity of high-quality, domain-specific metadata.

Apart from creating a new LLM, we have methods to augment existing LLMs for our task. One way of doing that is prompt engineering, which involves carefully designing the input given to a language model to elicit the desired output. This can include providing context by including relevant background information or instructions in the prompt. It can also involve using few-shot examples by providing a few examples of the task to guide the model's response. Additionally, structured prompts using templates or specific formatting can guide the model. For metadata-related tasks in bioinformatics, prompt engineering can involve specifying the structure of the metadata, including required fields, and providing examples of well-formed metadata entries. This information can help guide the model to generate more accurate and consistent metadata. Our proposed method is a form of prompt engineering which derives information from CEDAR for context.

Our investigation revealed that, while prompt engineering shows promise in leveraging domain knowledge, there might be scope for improvement by having a knowledge base for the task of prompting itself. In our experiment, while GPT-4 alone could make linguistic corrections, it could not ensure completeness, correctness, or consistency. For example, it could not by itself produce links to ontologies, a criterion for findability and interoperability, even after being explicitly prompted to do so. Adding the name of the required ontology, as specified by the CEDAR metadata template, enhanced the LLM's performance on this front. By leveraging CEDAR templates, we can tailor prompt engineering strategies to specific domains, thereby enhancing the language model's ability to adhere to community standards in diverse contexts. This objective aligns with the principles of the FAIR data initiative, emphasizing the need for high-quality metadata in ensuring that data are findable, accessible, interoperable, and reusable.

Our study also reveals that, although the combined usage of GPT-4 and CEDAR is powerful for metadata standardization, the process is still prone to a few errors. In our investigation, we found that the reviewers often disagree regarding possible metadata errors. We have included a diverse set of reviewers who score errors in the records (whether they may be adherence errors or incorrect transformation of the input BioSample record) differently. Given the variability of factors, the scoring varied among reviewers. However, the reviewers *consistently* score the GPT-4 and CEDAR records significantly higher than both GPT-4 augmented records and the original BioSample records.

Another popular method for enhancing LLM performance is to use structured knowledge. One way of doing this is to use a Retrieval Augmented Generation (RAG) pipeline. RAG pipelines[18] combine retrieval-based methods with generative models to enhance the text-generation process with relevant external knowledge. The pipeline first retrieves relevant documents or pieces of information from a large knowledge base. This retrieved information is then used to augment the input to a generative model (an LLM). The generative model, now enriched with relevant domain knowledge, produces the final output. For example, in bioinformatics, a RAG pipeline might retrieve relevant research papers, clinical trial results, or database entries to augment the generation of a scientific report or metadata description. This approach can significantly improve the accuracy and relevance of the generated content by grounding it in real-world data. In the future, we can experiment with multiple domains and use CEDAR as source for a RAG based architecture. Since we restricted ourselves to a single database, we used the corresponding CEDAR metadata template directly. A second way of incorporating knowledge is using knowledge graphs. A knowledge graph is a structured representation of knowledge that illustrates the relationships among different entities in some application area. In bioinformatics, knowledge graphs can be used to model complex biological systems, such as the interactions between genes, proteins, diseases, and drugs. By incorporating domain-specific knowledge, knowledge graphs can help to enhance the performance of language models in tasks such as information retrieval, question answering, and data integration. However, the semi-structured nature of metadata makes it challenging to directly apply knowledge graphs. Metadata often include a mix of structured information (such as key–value pairs) and unstructured text (such as prose descriptions), making it difficult to map everything into a graph format.

Our study also reveals that, although the combined usage of GPT-4 and CEDAR is powerful for metadata standardization, the process is still prone to a few errors. Our research highlights the potential of combining the strengths of large language models such as GPT-4 with structured knowledge sources to address complex data challenges, such as repairing messy, legacy metadata. By making existing online data "AI ready," this approach opens new avenues for leveraging AI technologies for data curation at scale. Additionally, efforts to expand the availability and accessibility of structured knowledge sources within CEDAR can help in realizing the full potential of language models in enhancing metadata.

While our experiments have demonstrated the efficacy of this approach, further research is warranted to explore its applicability across a broader range of datasets and domains. We also plan to extend the study to examine field-by-field improvement, rather than record-level improvement, for developing better insights. We also plan to explore the use of our approach to perform other metadata-related tasks, such as harmonization, extraction, and summarization. Implementing our approach to clean up all the records in BioSample, and extending it to all repositories at NCBI, would mark a substantial stride toward enhancing data integrity and accessibility within the scientific community. By systematically applying our methodology across these vast repositories, we could improve consistency and completeness in metadata descriptions, thereby fostering trust in the data and facilitating seamless collaboration and interoperability of scientific artifacts. Moreover, a scientific landscape where datasets adhere to the FAIR principles and are readily available for exploration and secondary analysis will be transformative. Such a scenario would democratize scientific knowledge, empowering researchers worldwide to conduct comprehensive analyses, uncover new insights, and accelerate biomedical discoveries.

**Conclusion**
Our experiments have shed light on the capabilities of including a structured knowledge base for metadata (the body of templates in CEDAR) along with GPT-4 for ensuring metadata adherence to community standards. Our findings underscore the challenges associated with applying GPT-4 to the task of enhancing metadata. The best adherence was recorded when GPT-4 was augmented with CEDAR, determined through both the automated and the peer-review evaluation. In automated evaluation, adherence for specific fields significantly improved from 40% to 77%. In peer-review evaluation, adherence for entire records improved from 79% to 97%. Similarly, the enforcement of generating the field name in a consistent and reproducible manner across hundreds of samples is ensured by the metadata template, a feature that is impossible to enforce with the variability of text in LLMs. The field values, further, are restricted to be from designated ontologies if specified so by CEDAR.

**Acknowledgments**
This work was supported in part by grant R01 LM013498 from the National Library of Medicine. We thank Jane Liang, Amy Zhang, Yingjie Weng, Anna Graber-Naidich and other members of the Qualitative Science Unit, School of Medicine at Stanford for their valuable input and suggestions for improving the study.

**Data and Code**
The data and code are available at https://github.com/musen-lab/BioSampleGPTCorrection.